\documentclass[10pt,conference]{IEEEtran}
\IEEEoverridecommandlockouts
\usepackage{cite}
\usepackage{amsmath,amssymb,amsfonts}
\usepackage{algorithmic}
\usepackage{graphicx}
\usepackage{textcomp}
\usepackage{xcolor}
\def\BibTeX{{\rm B\kern-.05em{\sc i\kern-.025em b}\kern-.08em
    T\kern-.1667em\lower.7ex\hbox{E}\kern-.125emX}}

\usepackage{hyperref}
\usepackage{url}
\usepackage{graphicx}
\usepackage{booktabs}
\usepackage{caption}
\usepackage{subfigure}
\usepackage{makecell}

\makeatletter
\renewcommand\normalsize{%
\@setfontsize\normalsize\@xpt\@xiipt
\abovedisplayskip 3\p@ \@plus3\p@ \@minus5\p@
\abovedisplayshortskip \z@ \@plus3\p@
\belowdisplayshortskip 6\p@ \@plus3\p@ \@minus3\p@
\belowdisplayskip \abovedisplayskip
\let\@listi\@listI}
\makeatother

\IEEEoverridecommandlockouts\IEEEpubid{\makebox[\columnwidth]{ 978-1-6654-3540-6/22/\$31.00~\copyright~2022 IEEE \hfill} \hspace{\columnsep}\makebox[\columnwidth]{ }}
\begin{document}

\title{FedDQ: Communication-Efficient Federated Learning with Descending Quantization
%\thanks{This work was put in arxiv: https://arxiv.org/abs/2110.02291.\\\indent{The code of this paper is available at https://github.com/lqu-001/FedDQ}}
}

\author{\IEEEauthorblockN{Linping Qu, Shenghui Song, Chi-Ying Tsui}
\IEEEauthorblockA{Dept. of ECE, The Hong Kong University of Science and Technology, Hong Kong\\
Email: lqu@connect.ust.hk, eeshsong@ust.hk, eetsui@ust.hk}
}

\maketitle

\begin{abstract}
Federated learning (FL) is an emerging learning paradigm without violating users' privacy. However, large model size and frequent model aggregation cause serious communication bottleneck for FL. To reduce the communication volume, techniques such as model compression and quantization have been proposed. Besides the fixed-bit quantization, existing adaptive quantization schemes use ascending-trend quantization, where the quantization level increases with the training stages. In this paper, we first investigate the impact of quantization on model convergence, and show that the optimal quantization level is directly related to the range of the model updates. Given the model is supposed to converge with the progress of the training, the range of the model updates will gradually shrink, indicating that the quantization level should decrease with the training stages. Based on the theoretical analysis, a descending quantization scheme named FedDQ is proposed. Experimental results show that the proposed descending quantization scheme can save up to 65.2\% of the communicated bit volume and up to 68\% of the communication rounds, when compared with existing schemes.
\end{abstract}


\section{Introduction}
Federated learning (FL) \cite{b1} is a distributed learning scheme that does not require sharing users' data, and hence can protect users' privacy. However, FL may face severe communication bottlenecks due to the frequent communication of large machine learning models between the server and the clients \cite{b2}. Many techniques have been proposed to tackle the communication obstacle, such as infrequent aggregation \cite{b3, b4}, sparse compression \cite{b5}, and quantization \cite{b6, b7}. The key idea of quantization is to use fewer bits to represent the model updates, which introduces a trade-off between the communication workload and representation accuracy. To this end, the selection of the quantization scheme becomes critical \cite{b8, b9}.

Fixed-bit quantization schemes, such as 8-bit quantization \cite{b6}, ternary gradients \cite{b7}, and even 1-bit quantization \cite{b10}, were proposed without considering the changing nature of the training process. Adaptive quantization was later proposed to take the training process into consideration. However, most proposed schemes adopted an ascending quantization with the intuition that fewer quantization bits, at early training stages, can save the communication volume and more quantization bits, at the later stages, can help achieve high accuracy. The adaptive scheme was designed by metrics including the root mean square value of gradients \cite{b8}, gradients' mean to standard deviation ratio \cite{b11}, and training loss \cite{b9}. In \cite{b12}, the adaptive quantization problem was theoretically investigated where the range of the model updates is approximated by a bound. 

In this paper, we first investigate the impact of quantization on convergence. For a given total communication volume, we optimize the allocation of the quantization bits to different training stages. In our derivation process, the range of model updates is not approximated to be a constant, but a variable instead. It is shown that the optimal quantization level exactly depends on the range of model updates, and the quantization level should have a decreasing relationship with the training stages. Based on the analysis, we propose a descending quantization strategy named FedDQ. Experimental results show that the proposed scheme can save both the total communicated bit volume and the number of communication round, when compared with the existing ascending quantization approaches.

The main contributions of this work are outlined as follows:\\
1) We show that the quantization level has a strong relation with the range of model updates.\\
2) Based on the theoretical result, we propose a descending-trend quantization scheme named FedDQ.\\
3) Experimental results demonstrate that FedDQ can reduce the communication volume, compared to the state-of-the-art ascending scheme, and also converges faster by consuming fewer communication rounds. 

\section{Preliminaries}

\subsection{Federated Learning}

The goal of FL is to train a global model $X\in {\mathbb{R}^d}$, where $d$ is the dimensions of the model, with multiple rounds of training on distributed datasets residing on different clients. The problem can be formulated as \cite{b1, b13}
\begin{align}
{\underset{X}{min} f(X) := \underset{X}{min}\sum\nolimits_{i=1}^np_if_i(X)},
\end{align}
where $n$ is the number of clients, $p_i$ is the  ratio between the data size of the $i^{th}$ client and that of all clients, and $f_i(X)$ is the loss function for the $i^{th}$ client. In the $m^{th}$ round of communication, the server broadcasts the global model $X_m$ to all the selected clients, which will perform $\tau$ steps of local stochastic gradient descend (SGD) with a step size $\eta$. The local updating rule for the $i^{th}$ client is given by
\begin{align}
{X^{i}_{m,t+1}=X^{i}_{m,t}-\eta_{m,t}\tilde{\nabla} f_i(X^{i}_{m,t})},
\end{align}
where $t=0,...,\tau-1$, $X^{i}_{m,0}=X_m$, and $\tilde{\nabla} f_i(X^{i}_{m,t})$ denotes the stochastic gradient computed from local datasets. After completing $\tau$ steps of local training, the $i^{th}$ client obtains a new model $X^{i}_{m,\tau}$, and the model update of the $i^{th}$ client is calculated by 
\begin{align}
{\Delta X^{i}_m=X^{i}_{m,\tau}-X_m}.
\end{align}
Before uploading the model update $\Delta X^{i}_m$ to the server, quantization is applied to get $Q(\Delta X^{i}_m)$. At the server, the uploaded parameters from $r$ selected clients will be aggregated to update the global model by
\begin{align}
X_{m+1}=X_m+\frac{1}{r}\sum_{i\in{S_m}}Q(\Delta X^{i}_m),
\end{align}
where $S_m$ is the set of selected clients.

\subsection{Stochastic Uniform Quantization}
In this paper, we adopt a common quantizer named a stochastic uniform quantizer \cite{b14}. For a given model update $\Delta X_i \in {\mathbb{R}^d}$, we first compute its range, i.e., $range(\Delta X_i) = \Delta X_i^{max}- \Delta X_i^{min}$, where $\Delta X_i^{max}=max_{1\leq j\leq d}\Delta X_i(j)$, $\Delta X_i^{min}=min_{1\leq j\leq d}\Delta X_i(j)$ denote the maximum and minimum values of all the model updates, respectively.
With $N$ bit quantization, the range is divided into $2^N-1$ bins. The quantized value of $\Delta X_i(j)$ depends on which bin it is located in. For example, if $\Delta X_i(j)$ is located in one bin whose lower bound and upper bound are $h'$ and $h''$, respectively, then the quantized value is given by 
\begin{eqnarray}
Q(\Delta X_i(j))=
\begin{cases}
h', & \mbox{with probability  }\frac{h''-\Delta X_i(j)}{h''-h'}, \\
h'', & \mbox{otherwise}.
\end{cases}
\end{eqnarray}

\section{Intuitive and Theoretical Analysis}
\label{Intuitive and Theoretical Analysis}
In this section, we first discuss the drawbacks of the existing ascending quantization schemes intuitively. Then, we formulate the quantization problem and analyze it theoretically.  
\subsection{Intuitive Analysis}
\begin{figure}[t]
\begin{minipage}[b]{0.48\linewidth}
  \centering
  \centerline{\includegraphics[width=3.8cm]{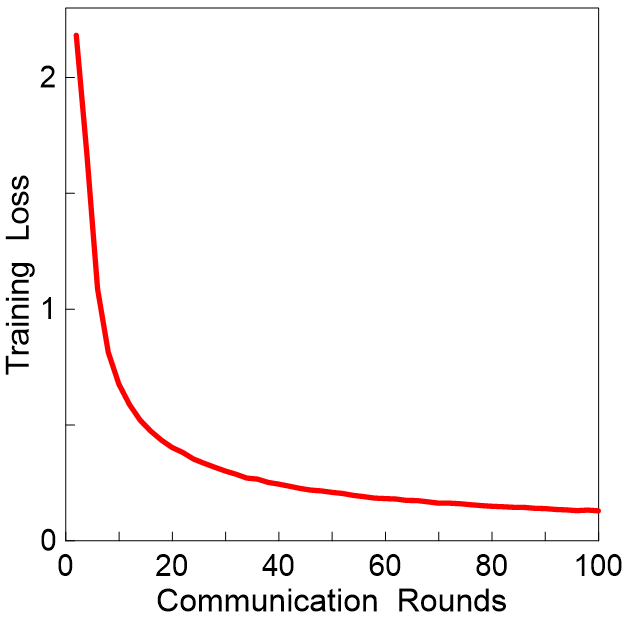}}
%  \vspace{1.5cm}
  \centerline{(a) Training loss curve.}
  %\medskip
\end{minipage}
\hfill
\begin{minipage}[b]{0.48\linewidth}
  \centering
  \centerline{\includegraphics[width=4.0cm]{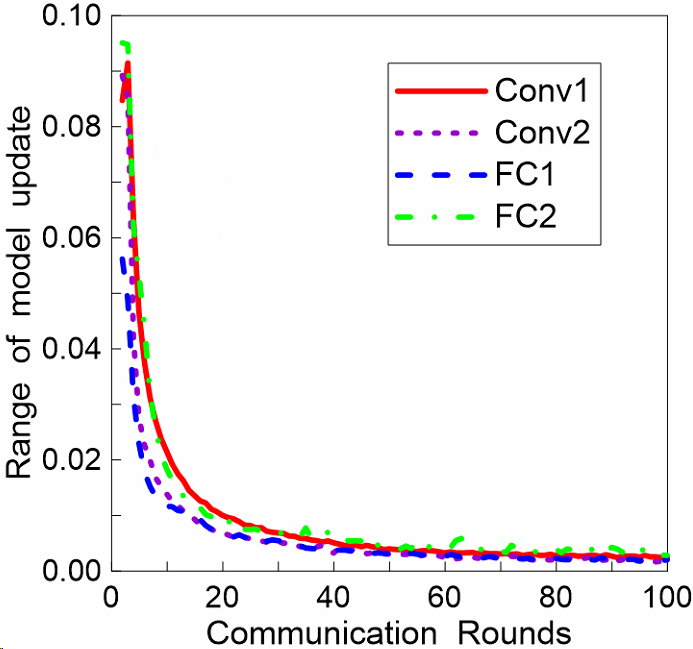}}
%  \vspace{1.5cm}
  \centerline{(b) Model update range curve.}
  %\medskip
\end{minipage}
\caption{The training characteristics which show the decreasing training loss and decreasing range of model updates. This example is FL on Fahion-MNIST\cite{b15} using Vanilla CNN\cite{b1} model.}
\label{fig:char}
\end{figure}

Existing adaptive quantization strategies use an ascending quantization based on the intuition that a low quantization level at the early training stages can save communication volume, while a high quantization level at the late stages can improve the training convergence \cite{b11}. However, this intuition may not reflect the actual situation for the following reasons:

1) In the early training stages, the training loss drops very quickly (as shown in Fig.~\ref{fig:char}(a)), and a high quantization level at the early stages can enhance the speed of this drop. Using a low quantization level at the early stages does save the bit volume, but may slow down the convergence, which results in more communication rounds and larger overall communication volume to reach convergence.

2) In the later training stages, the model starts to converge and becomes stable. As a result, the range of model updates for each layer will decrease (as shown in Fig.~\ref{fig:char}(b)). Thus, a small bit-length is enough to represent the narrower range, while using a high number of quantization bits will be a waste.

Based on the above reasons, a descending-trend quantization is more suitable for FL training, and we support this intuition with theoretical analysis in the following section. 
\subsection{Theoretical Analysis}
In this section, we first derive the convergence bound for a given communication volume constraint. Then, we determine the optimum quantization level by maximizing the convergence rate. We begin by introducing three commonly-used assumptions.\\
\textbf{Assumption 1.} \textit{The random quantizer Q($\cdot$) is unbiased and its variance is bounded by the quantization level and the range of the parameters, i.e.,} $\mathbb{E}[Q(X)|X]=X$ $and$ $\mathbb{E}[||Q(X)-X||^2|X]\leq q_s(range(X))^2$  \cite{b14}, $where$ $q_s=\frac{d}{s^2}$ \cite{b12} \textit{and s is the number of quantization bins, i.e., range(X) is divided into s bins.}\\
\textbf{Assumption 2.} \textit{The loss function $f_i$ is L-smooth with respect to X, i.e., for any X, $\hat{X} \in \mathbb{R}^d$, we have $||\nabla f_i(X)-\nabla f_i(\hat{X})||\leq L||X-\hat{X}||$}\cite{b12, b13}.\\
\textbf{Assumption 3.} \textit{The stochastic gradient $\tilde{\nabla} f_i(X)$ is unbiased and variance bounded, i.e., $\mathbb{E}_{\xi}[\tilde{\nabla} f_i(X)]=\nabla f_i(X)$, and $\mathbb{E}_{\xi}[||\tilde{\nabla} f_i(X)-\nabla f_i(X)||^2] \leq \sigma^2, \xi$ is the mini-batch dataset.}

With Assumptions 1-3, the convergence bound of FL can be given by the following theorem.\\
\noindent\textbf{Theorem 1.} \textit{ For given K communication rounds, the convergence is bounded by \\
\begin{align}
&\frac{1}{K\tau}\sum_{m=0}^{K-1}\sum_{t=0}^{\tau -1}\mathbb{E}\Vert \nabla f(\bar{X}_{m,t})\Vert^2 \leq \frac{Ld}{n^2\eta K\tau}\sum_{m=0}^{K-1}\sum_{i\in [n]}(\frac{range_m^{i}}{s_m^{i}})^2 \nonumber \\
&+\frac{2(f(X_0)-f^*)}{\eta K\tau}+\frac{\eta^2\sigma^2(n+1)(\tau-1)L^2}{n} +\frac{\eta \sigma^2L}{n}, 
\label{Theorem}
\end{align}
where $\bar{X}_{m,t}$ is the averaged model on all local clients, $range_m^i$ is the range of model updates for the $i^{th}$ client in the $m^{th}$ communication round, $f^*$ is the minimum value of training loss, and the other symbols were defined previously. Proof of the theorem is provided in the Appendix.}

Our target is to minimize the right hand side of (\ref{Theorem}). With the constraint of a total communication volume $B=\sum_{m=0}^{K-1}\sum_{i\in [n]}ds^i_m$, we want to optimize $s^i_m$ to minimize the right hand side of (\ref{Theorem}). By ignoring the three terms on the most right hand side of (\ref{Theorem}), the problem can be simplified as 
\begin{align}
\mathop{min}\limits_{s^i_m} \ \sum_{m=0}^{K-1}\sum_{i\in [n]}(\frac{range_m^{i}}{s_m^{i}})^2 \nonumber\\
s.t. \quad \sum_{m=0}^{K-1}\sum_{i\in [n]}ds^i_m = B.
\label{optimization}
\end{align}

To solve above optimization problem, we recall the Cauchy-Schwarz inequality: $(\sum_{i=1}^nx_i^2)(\sum_{i=1}^ny_i^2) \geq (\sum_{i=1}^nx_iy_i)^2$, where the equality holds when $x_1/y_1 = x_2/y_2=...=x_n/y_n$. By setting $y_i=1$, we get the variant of Cauchy-Schwarz inequality: $(\sum_{i=1}^nx_i^2) \geq 1/n(\sum_{i=1}^nx_i)^2$, and the equality holds when $x_1 = x_2=...=x_n$.

Applying the variant of Cauchy-Schwarz inequality on (\ref{optimization}), we get
\begin{align}
\sum_{m=0}^{K-1}\sum_{i\in [n]}(\frac{range_m^{i}}{s_m^{i}})^2 \geq \sum_{m=0}^{K-1}\frac{1}{n}(\sum_{i\in [n]}\frac{range_m^{i}}{s_m^{i}})^2,
\label{Cauchy1}
\end{align}
and the condition for equality is 
\begin{align}
\frac{range^0_m}{s^0_m}=\frac{range^1_m}{s^1_m}=...=\frac{range^n_m}{s^n_m}.
\label{condition1}
\end{align}
Continuing to apply the variant of Cauchy-Schwarz inequality on (\ref{Cauchy1}), we get
\begin{align}
\sum_{m=0}^{K-1}\frac{1}{n}(\sum_{i\in [n]}\frac{range_m^{i}}{s_m^{i}})^2 \geq \frac{1}{Kn}(\sum_{m=0}^{K-1}\sum_{i\in [n]}\frac{range_m^{i}}{s_m^{i}})^2,
\label{Cauchy2}
\end{align}
and the condition for equality is 
\begin{align}
\sum_{i\in [n]}\frac{range^i_0}{s^i_0}=\sum_{i\in [n]}\frac{range^i_1}{s^i_1}=...=\sum_{i\in [n]}\frac{range^i_{K-1}}{s^i_{K-1}}.
\label{condition2}
\end{align}
Combining (\ref{Cauchy1}) and (\ref{Cauchy2}), we get
\begin{align}
\sum_{m=0}^{K-1}\sum_{i\in [n]}(\frac{range_m^{i}}{s_m^{i}})^2 \geq \frac{1}{Kn}(\sum_{m=0}^{K-1}\sum_{i\in [n]}\frac{range_m^{i}}{s_m^{i}})^2,
\label{Cauchy}
\end{align}
and the condition for equality is, for $i = 0,1,...,n$ and $m=0,1,...,K-1$, 
\begin{align}
\frac{range^i_m}{s^i_m}=\alpha,
\label{condition}
\end{align}
 where $\alpha$ is a constant. From (\ref{optimization}) we can get $\alpha=\frac{d}{B}\sum_{m=0}^{K-1}\sum_{i\in [n]}range^i_m$.

Thus, the optimized quantization bins for the $i^{th}$ client in the $m^{th}$ communication round is 
\begin{align}
s^i_m = \frac{range^i_m}{\alpha}.
\label{bin}
\end{align}
It shows that the optimal quantization bins should be proportional to the model update range. From Fig.~\ref{fig:char}(b), we know that $rang_m^i$ will shrink with communication round $m$. This supports the use of a descending quantization scheme.

\section{Design and Convergence proof}
\subsection{Design of Descending Quantization}
Eq.(\ref{bin}) indicates that the optimized quantization bins should be proportional to its model update range. But the constant $\alpha=\frac{d}{B}\sum_{m=0}^{K-1}\sum_{i\in [n]}range^i_m$ is difficult to determine because $B$ is not same for all experiments, and we also don't know the range for the further communication rounds. So we keep $\alpha$ as a hyper-parameter.

For the $i^{th}$ client in the $m^{th}$ communication round, the quantization scheme is
\begin{align}
&s^i_m = \frac{range^i_m}{\alpha} \nonumber\\
&bit_m^i= \lceil log_2(s^i_m) \rceil.
\label{finalbit}
\end{align}
The quantization bit will only change with the variable $range_m^i$, which has a descending characteristic.

Eq.(\ref{finalbit}) is practical and easy to use in reality. The client just needs to compute the range of local updates and then decide the quantization bit-length. Compared with other quantization schemes, FedDQ has more freedom as each client can decide the quantization bit individually.

\subsection{Convergence Proof}
We now provide the convergence proof for FedDQ in non-convex scenarios. By substituting $s^i_m=range^i_m/\alpha$ into (\ref{Theorem}), and setting the learning stepsize as $\eta=1/L\sqrt{K\tau}$ \cite{b13}, then the following first-order stationary condition holds:
\begin{align}
\frac{1}{K\tau}\sum_{m=0}^{K-1}\sum_{t=0}^{\tau -1}\mathbb{E}\Vert \nabla &f(\bar{X}_{m,t})\Vert^2 \leq \frac{2L(f(X_0)-f^*)}{\sqrt{K\tau}}\nonumber\\
+\frac{L^2d\alpha^2}{n\sqrt{K\tau}} 
&+\frac{\sigma^2}{n\sqrt{K\tau}}
+\frac{\sigma^2(n+1)(\tau-1)}{nK\tau}
. \label{rate}
\end{align}
The result implies the following convergence rate:
\begin{align}
\frac{1}{K\tau}\sum_{m=0}^{K-1}\sum_{t=0}^{\tau -1}\mathbb{E}\Vert \nabla f(\bar{X}_{m,t})\Vert^2 \leq \mathcal{O}(\frac{1}{\sqrt{K\tau}}). \label{rate1}
\end{align}
Eq.(\ref{rate1}) shows if communication rounds $K \to \infty$, we have $\frac{1}{K\tau}\sum_{m=0}^{K-1}\sum_{t=0}^{\tau -1}\mathbb{E}\Vert \nabla f(\bar{X}_{m,t})\Vert^2 \to 0$. It proves that the FL training will converge with sufficient number of communication rounds $K$.

\section{Experiments and Discussions}
\subsection{Experiment Setup}
We compare FedDQ with the state-of-the-art ascending quantization scheme AdaQuantFL \cite{b12} to show the effectiveness of our approach. As \cite{b12} already shows that adaptive quantization performs better than fixed-bit quantization, we will not compare FedDQ with fixed quantization schemes. Three benchmarks are used in the experiments: 1) Vanilla CNN \cite{b1} on Fashion MNIST \cite{b15}, 2) CNN (4 convolution layers + 3 fully connected layers) on CIFAR-10 \cite{b16}, and 3) ResNet-18 \cite{b17} on CIFAR-10. 

We use a similar experimental setup as that used for AdaQuantFL in \cite{b12}. The training datasets of Fashion MNIST and CIFAR-10 are split among all clients, and the test datasets are used to perform validation on the server side. For the  hyper-parameters, we set the local update steps $\tau$=5, the step size $\eta$=0.1, and $\alpha=0.005$. An SGD optimizer is utilized for the local training. The numbers of local clients for the three benchmarks are 10, 10, and 4, respectively. NVIDIA 3090 GPU, CUDA 11.4, and PyTorch 0.2.2 are deployed in our experiments.

\subsection{Experiment Results}
\begin{figure}[t]
\begin{minipage}[b]{.48\linewidth}
  \centering
  \centerline{\includegraphics[width=4.0cm]{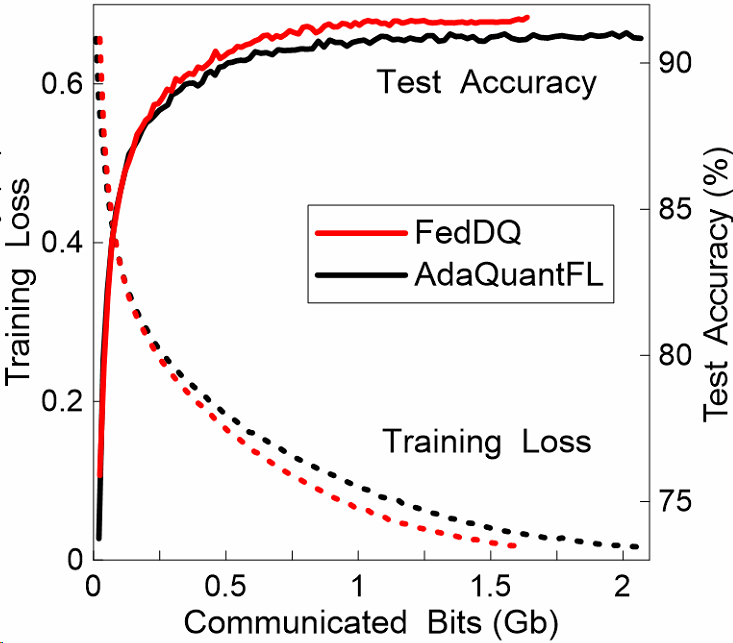}}
%  \vspace{1.5cm}
  \begin{center}
  (a) Performance vs.\\
  communicated bits.
  \end{center}
  %\medskip
\end{minipage}
\hfill
\begin{minipage}[b]{0.48\linewidth}
  \centering
  \centerline{\includegraphics[width=4.0cm]{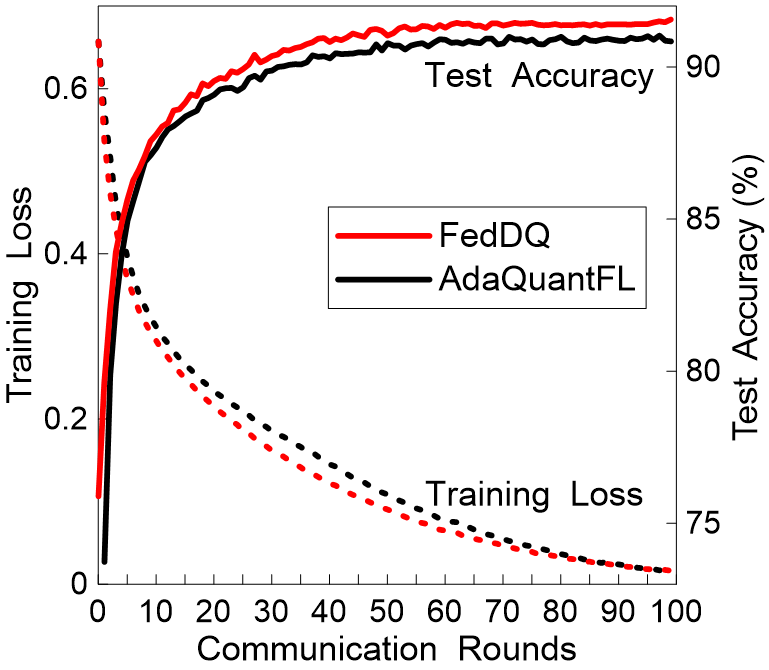}}
%  \vspace{1.5cm}
  \begin{center}
  (b) Performance vs.\\
  communicated rounds.
  \end{center}
  %\medskip
\end{minipage}
\caption{Vanilla CNN on Fashion MNIST.}
\label{fig:vanilla}
\end{figure}

\begin{figure}[t]
\begin{minipage}[b]{.48\linewidth}
  \centering
  \centerline{\includegraphics[width=4.0cm]{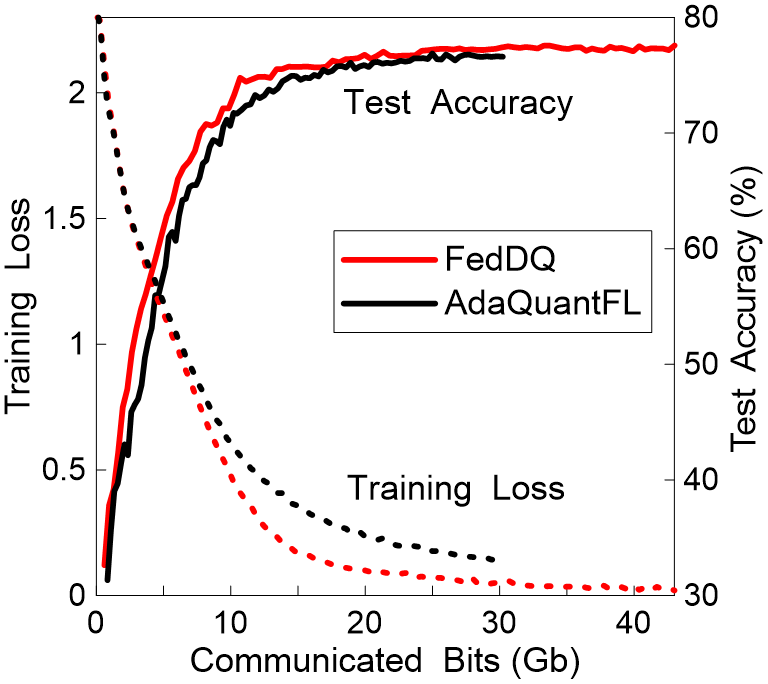}}
%  \vspace{1.5cm}
  \begin{center}
  (a) Performance vs.\\
  communicated bits.
  \end{center}
  %\medskip
\end{minipage}
\hfill
\begin{minipage}[b]{0.48\linewidth}
  \centering
  \centerline{\includegraphics[width=4.0cm]{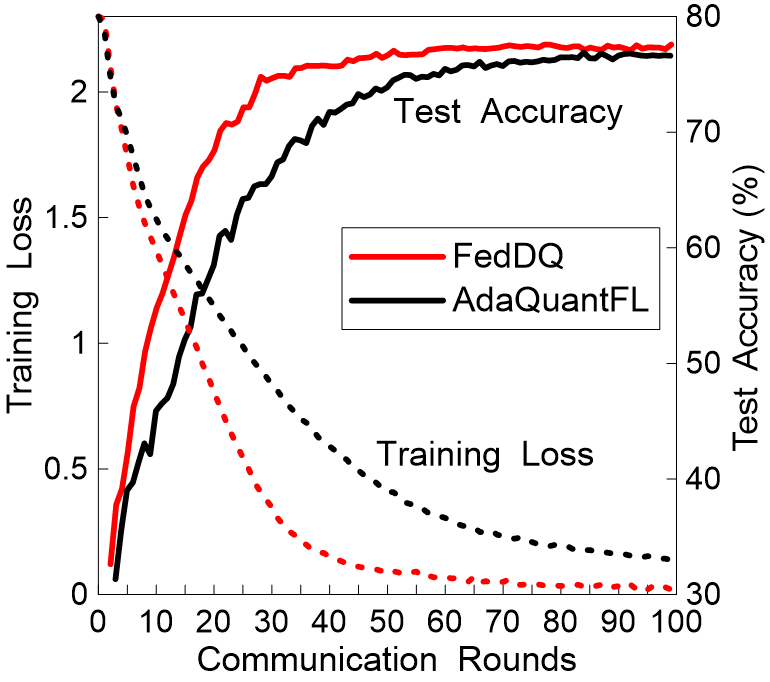}}
%  \vspace{1.5cm}
  (b) Performance vs.\\
  communicated rounds.
  %\medskip
\end{minipage}
\caption{CNN on CIFAR-10.}
\label{fig:cnn}
\end{figure}

\begin{figure}[t]
\begin{minipage}[b]{.48\linewidth}
  \centering
  \centerline{\includegraphics[width=4.0cm]{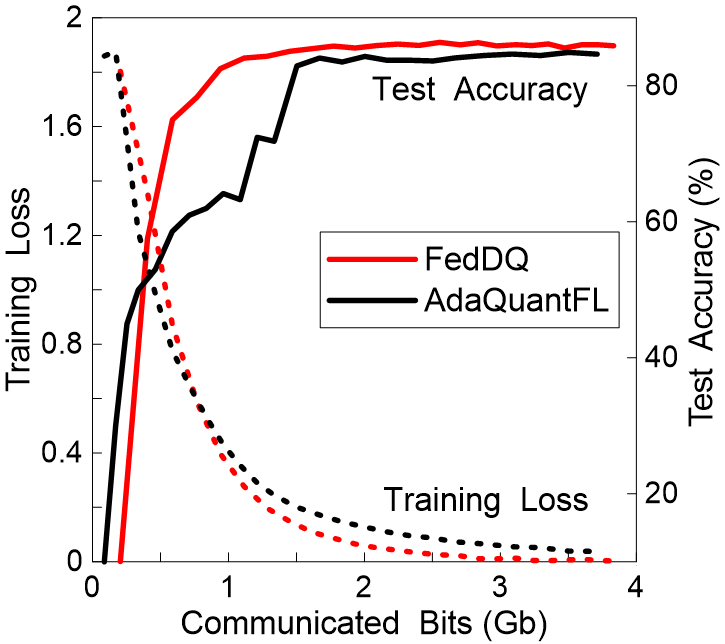}}
%  \vspace{1.5cm}
  \begin{center}
  (a) Performance vs.\\
  communicated bits.
  \end{center}
  %\medskip
\end{minipage}
\hfill
\begin{minipage}[b]{0.48\linewidth}
  \centering
  \centerline{\includegraphics[width=4.0cm]{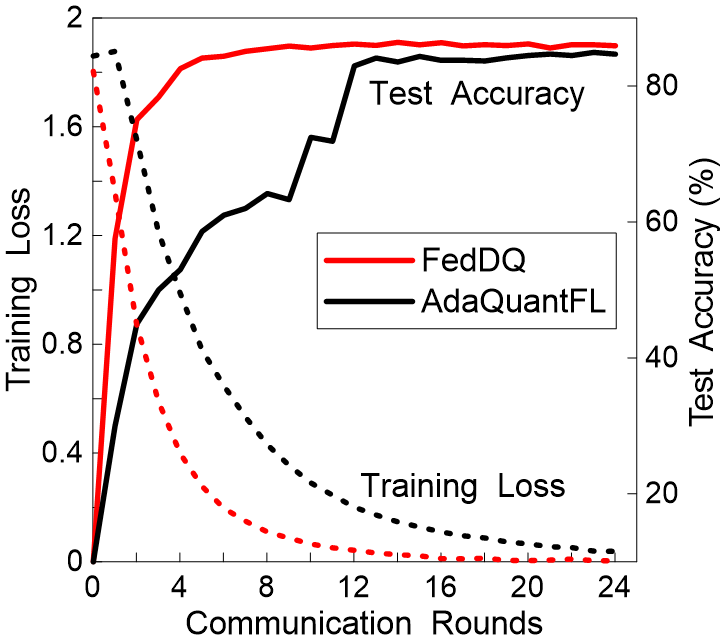}}
%  \vspace{1.5cm}
  (b) Performance vs.\\
  communicated rounds.
  %\medskip
\end{minipage}
\caption{ResNet-18 on CIFAR-10.}
\label{fig:resnet}
\end{figure}

Both the training loss and test accuracy are checked for all the experiments. The training loss is the averaged value of all the clients' local training loss \cite{b18}. We also compare the total communication volume and the number of communication rounds for both methods. The comparisons are shown in Figures \ref{fig:vanilla} to \ref{fig:resnet}.

In Fig.~\ref{fig:vanilla}(a), Fig.~\ref{fig:cnn}(a) and Fig.~\ref{fig:resnet}(a), we compare the performance with respect to the total communication volume. We can see that to achieve the same test accuracy or the same training loss, the number of bits that need to be transmitted for the proposed FedDQ scheme is fewer than that of AdaQuantFL. This is because, in the later stages of training, the model range becomes smaller, though AdaQuantFL keeps increasing the quantization bits, which may cause bit volume waste. On the other hand, FedDQ uses smaller quantization bits while maintaining the accuracy.

In Fig.~\ref{fig:vanilla}(b), Fig.~\ref{fig:cnn}(b) and Fig.~\ref{fig:resnet}(b), we compare the performance regarding the communication rounds. Similarly to the case of the bit volume, to achieve the same test accuracy or the same training loss, the proposed FedDQ scheme requires fewer communication rounds. This is because, in the early training stages, the training loss drops very quickly. AdaQuantFL uses small quantization bit-length in these stages, which may slow down the convergence speed and increases the overall number of communication rounds. FedDQ assigns higher quantization bit-length in these stages to accelerate the convergence and hence, it can reduce the number of communication rounds and converge faster.
 \begin{table}[b]
\caption{Performance Improvement}
\centering
\begin{tabular}{c|c|c|c}
\hline  
%x&Communicated bits&Communication rounds&x&x \\
&\multicolumn{3}{c}{Communicated Bits} \\
\hline  
&\makecell[c]{AdaQuantFL}&\makecell[c]{FedDQ}&\makecell[c]{Reduction Ratio}\\
\hline  
\makecell[c]{Experiment.1\\ Acc.=91.0\%}&2.07 Gb&\makecell[c]{0.72 Gb}&\makecell[c]{65.2\%}\\
\hline  
\makecell[c]{Experiment.2\\ Acc.=76.7\%}&30.25 Gb&\makecell[c]{19.95 Gb}&\makecell[c]{34.0\%}\\
\hline  
\makecell[c]{Experiment.3\\Acc.=84.7\%}&3.71 Gb&\makecell[c]{1.45 Gb}&\makecell[c]{60.9\%}\\

\hline
%x&Communicated bits&Communication rounds&x&x \\
&\multicolumn{3}{|c}{Communication Rounds} \\
\hline  
&\makecell[c]{AdaQuantFL}&\makecell[c]{FedDQ}&\makecell[c]{Reduction Ratio}\\
\hline  
\makecell[c]{Experiment.1\\Acc.=91.0\%}&100&\makecell[c]{43}&\makecell[c]{57\%}\\
\hline  
\makecell[c]{Experiment.2\\Acc.=76.7\%}&100&\makecell[c]{49}&\makecell[c]{51\%}\\
\hline  
\makecell[c]{Experiment.3\\Acc.=84.7\%}&25&\makecell[c]{8}&\makecell[c]{68\%}\\
\hline
\end{tabular}
%\caption{Performance improvement}
\label{tab:per}
\end{table}
 
 \begin{figure}[t]
\begin{minipage}[b]{0.48\linewidth}
  \centering
  \centerline{\includegraphics[width=4.0cm]{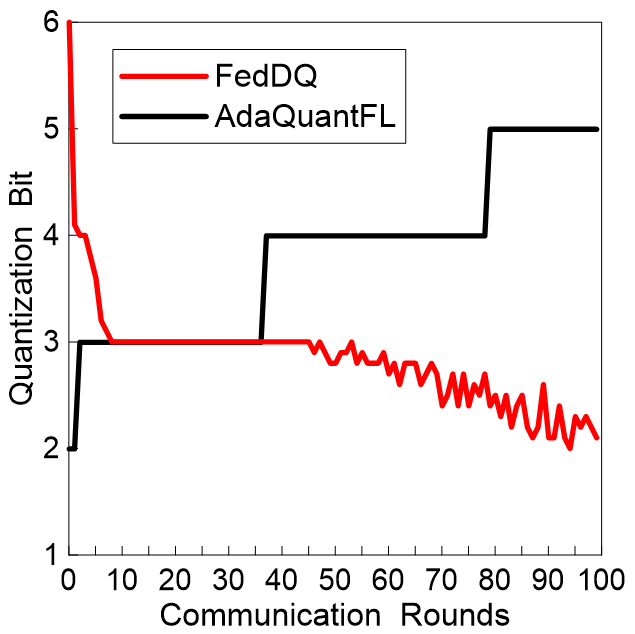}}
%  \vspace{1.5cm}
  \begin{center}
  (a) Vanilla CNN on \\Fashion MNIST.
  \end{center}
  %\medskip
\end{minipage}
\hfill
\begin{minipage}[b]{0.48\linewidth}
  \centering
  \centerline{\includegraphics[width=4.0cm]{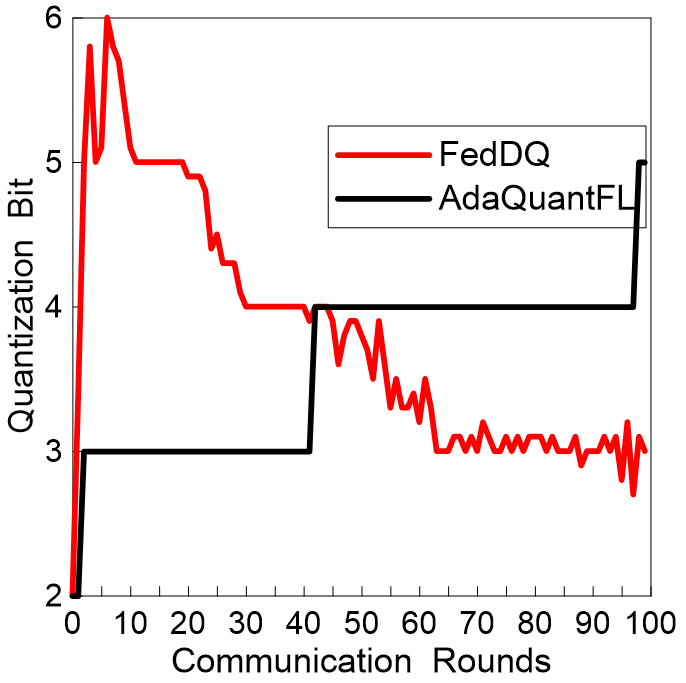}}
%  \vspace{1.5cm}
  \begin{center}
  (b) CNN on \\CIFAR-10.
  \end{center}
  %\medskip
\end{minipage}
\caption{Bit-length change curves in different experiments.}
\label{fig:bit}
\end{figure}

Table~\ref{tab:per} summarizes the comparison with respect to the consumed communication volume and communication rounds between FedDQ and AdaQuantFL. For example, in experiment 1, to achieve a test accuracy of $91.0\%$, AdaQuantFL consumes a total 2.07Gb in 100 rounds, while the proposed FedDQ only needs 0.72Gb in 43 rounds, which translates into a $65.2\%$ reduction in communication volume and $57\%$ reduction in the number of communication rounds.
 
Fig.~\ref{fig:bit} shows how the quantization bit-length  changes with the training stages for different methods. It can be observed that the proposed FedDQ results in a descending-trend quantization, while AdaQuantFL shows an ascending-trend quantization. In FedDQ, the quantization bit-length can be different for different clients, and the quantization bit-length shown in Fig.~\ref{fig:bit} is the average among all clients and thus, it is not an integer.
\section{Conclusion}
In this paper, we investigated adaptive quantization schemes for federated learning, and determined the optimal quantization level by maximizing the convergence rate. Theoretical analysis indicates that the quantization level should be related to the range of model updates, which shows a descending trend. Based on this result, we proposed an adaptive quantization scheme where the quantization level decreases with the training stages. A convergence guarantee was provided. Experimental results demonstrated that to achieve the same training loss or test accuracy, the proposed scheme requires a lower communicated bit volume and converges faster than the scheme with an ascending-trend quantization. 

\section*{Appendix}
We first define some notations which will be used throughout the proof. For each communication round $m=0,1,..., K-1$ and local iteration $t=0,1,...,\tau-1$, we denote
\begin{align}
&X_{m+1}=X_m+\frac{1}{r}\sum_{i\in{S_m}}Q(X_{m,\tau}^{i}-X_m),\nonumber\\
&\hat{X}_{m+1}=X_m+\frac{1}{n}\sum_{i\in{[n]}}Q(X_{m,\tau}^{i}-X_m),\nonumber\\
&\bar{X}_{m,t}=\frac{1}{n}\sum_{i\in{[n]}}X_{m,t}^{i},
\label{add_notation}
\end{align}
where $X_{m+1}$ is the updated global model based on $r$ selected clients, $\hat{X}_{m+1}$ is the updated global model based on all $n$ clients, and $\bar{X}_{m,t}$ is the averaged model of all $n$ raw clients without quantization.\\
\textbf{Lemma 1.} Consider the sequence of update ${X_{k+1},\hat{X}_{k+1},\bar{X}_{k,\tau}}$. If Assumptions 1 and 2 hold, then we have
\begin{align}
\mathbb{E}f(X_{m+1})&\leq \mathbb{E}f(\bar{X}_{m,\tau})+\frac{L}{2}\mathbb{E}\Vert \hat{X}_{m+1}-\bar{X}_{m,\tau}\Vert^2 \nonumber\\
&+\frac{L}{2}\mathbb{E}\Vert \hat{X}_{m+1}-X_{m+1}\Vert^2. 
\label{lemma1}
\end{align}
The proof is given in Section 8.2 of\cite{b13}.
In following three lemmas, we determine the upper bound for the three terms in the right-hand side (RHS) of (\ref{lemma1}).\\
\textbf{Lemma 2.} Given Assumptions 2 and 3, and considering the sequence of updates with stepsize $\eta$. Then we have 
\begin{align}
&\mathbb{E}f(\bar{X}_{m,\tau}) \leq \mathbb{E}f(X_m)-\frac{\eta}{2}\sum_{t=0}^{\tau -1}\mathbb{E}\Vert \nabla f(\bar{X}_{m,t})\Vert^2 \nonumber\\
&-\eta(\frac{1}{2n}-\frac{L\eta}{2n}-\frac{L^2\eta^2\tau(\tau-1)}{n})\sum_{t=0}^{\tau -1}\sum_{i \in [n]}\mathbb{E}\Vert \nabla f(X_{m,t}^{i})\Vert^2 \nonumber\\
&+\frac{L\tau\eta^2\sigma^2}{2n}+\frac{L^2\eta^3\sigma^2(n+1)\tau(\tau-1)}{2n}.
\end{align}
The proof can be found in \cite{b13}, Section 8.3. This Lemma shows that by receiving the global model $X_m$, each client will execute $\tau$ steps of local update, and after the local update, the training loss of the averaged clients' model $\bar{X}_{m,\tau}$ will be smaller than the loss of the original global model $X_m$. \\
\textbf{Lemma 3.} With Assumption 1, for sequences ${\hat{X}_{m+1},\bar{X}_{m,\tau}}$ defined in (\ref{add_notation}), we have 
\begin{align}
\mathbb{E}\Vert \hat{X}_{m+1}-\bar{X}_{m,\tau}\Vert^2 \leq \frac{1}{n^2}\sum_{i\in [n]}q_s(range_m^{i})^2,
\end{align}
where $q_s=\frac{d}{s^2}$.\\
Proof:
According to the definitions in (\ref{add_notation}), $\hat{X}_{m+1}$ and $\bar{X}_{m,\tau}$ are both the updated global model based on all clients, the difference is that $\hat{X}_{m+1}$ has up-link quantization error while $\bar{X}_{m,\tau}$ does not. Here we calculate the distance between these two models. Using Assumption 1, we have
\begin{align}
\mathbb{E}
&\Vert \hat{X}_{m+1}- \bar{X}_{m,\tau} \Vert^2\nonumber\\
&= \mathbb{E}\Vert X_m+\frac{1}{n}\sum_{i \in [n]}Q(X_{m,\tau}^i-X_m)  - \frac{1}{n}\sum_{i \in [n]}X_{m,\tau}^i \Vert^2 \nonumber\\
&= \mathbb{E}\Vert \frac{1}{n}\sum_{i \in [n]}Q(X_{m,\tau}^i-X_m)  - \frac{1}{n}\sum_{i \in [n]}(X_{m,\tau}^i-X_m) \Vert^2 \nonumber\\
&= \frac{1}{n^2}\mathbb{E}\Vert \sum_{i \in [n]}Q(X_{m,\tau}^i-X_m)  - \sum_{i \in [n]}(X_{m,\tau}^i-X_m)\Vert^2 \nonumber\\
&= \frac{1}{n^2}\mathbb{E}\Vert \sum_{i \in [n]}Q(\Delta X_m^i)  - \sum_{i \in [n]}\Delta X_m^i\Vert^2 \nonumber\\
&\leq \frac{1}{n^2}\sum_{i \in [n]}q_s(range_m^{i})^2.
\end{align}
\textbf{Lemma 4.} If Assumption 1 and 3 hold, then for the sequence of averages ${\hat{X}_{m+1}}$ defined in (\ref{add_notation}), we have 
\begin{align}
\mathbb{E}\Vert \hat{X}_{m+1}-&X_{m+1}\Vert^2 \leq \frac{4(1+q_s)}{r(n-1)}(1-\frac{r}{n})* \nonumber\\
&(n\sigma^2\tau\eta^2+\tau\eta^2\sum_{i\in [n]}\sum_{t=0}^{\tau-1}\Vert \nabla f(X_{m,t}^{i})\Vert^2).
\end{align}
The proof can be found in \cite{b13}, Section 8.5. It is to calculate the error by client selection. $\hat{X}_{m+1}$ is the updated global model based on all $n$ clients, $X_{m+1}$ is the updated global model based on the selected $r$ clients.

Now we get the basic lemmas in above 4 lemmas, we then continue to calculate the convergence bound. By combining Lemmas 1$\sim$4, we can get the following recursive inequality on the expected function value on the updated model at cloud server, i.e., ${X_m:m=1,...,K}$:
\begin{align}
&\mathbb{E}f(X_{m+1}) \leq \mathbb{E}f(X_m)-\frac{\eta}{2}\sum_{t=0}^{\tau -1}\mathbb{E}\Vert \nabla f(\bar{X}_{m,t})\Vert^2 \nonumber\\ &+\frac{L}{2n^2}\sum_{i\in [n]}q_s( range_m^i)^2-\frac{\eta}{2n}* \nonumber\\
&\left\{1-L\eta(1+\frac{4n(n-r)(1+q_s)\tau}{rn(n-1)})-2L^2\tau(\tau-1)\eta^2\right\} \nonumber\\
&*\sum_{t=0}^{\tau -1}\sum_{i \in [n]}\mathbb{E}\Vert \nabla f(X_{m,t}^{i})\Vert^2+\frac{L^2\eta^3\sigma^2(n+1)\tau(\tau-1)}{2n} \nonumber\\
& +\frac{L\eta^2\sigma^2\tau}{2}(\frac{1}{n}+\frac{4(1+q_s)n(n-r)}{rn(n-1)}).
\end{align}\\
When $\eta$ is small such that
\begin{align}
1-L\eta[1+\frac{1}{r(n-1)}&(1-\frac{r}{n})4n(1+q_s)\tau] \nonumber\\
&-2L^2\tau(\tau-1)\eta^2 \geq 0,
\label{smalleta}
\end{align}
we have
\begin{align}
\mathbb{E}&f(X_{m+1})\leq \mathbb{E}f(X_m)-\frac{1}{2}\eta\sum_{t=0}^{\tau -1}\mathbb{E}\Vert \nabla f(\bar{X}_{m,t})\Vert^2 \nonumber \nonumber\\
&+\frac{L}{2n^2}\sum_{i\in [n]}q_s( range_m^{i})^2+\eta^3\frac{\sigma^2}{n}(n+1)\frac{\tau(\tau-1)}{2}L^2 \nonumber\\ &+\eta^2\frac{L}{2}\sigma^2\tau[\frac{1}{n}+\frac{1}{r(n-1)}(1-\frac{r}{n})4(1+q_s)n]. \label{tosum}
\end{align}\\
Summing (\ref{tosum}) over $m=0,...,K-1$ and rearranging the terms produces that
\begin{equation}
\begin{split}
\frac{1}{2}\eta&\sum_{m=0}^{K-1}\sum_{t=0}^{\tau -1}\mathbb{E}\Vert \nabla f(\bar{X}_{m,t})\Vert^2 \leq f(X_0)-f^*\\
&+\frac{L}{2n^2}\sum_{m=0}^{K-1}\sum_{i\in [n]}q_s(range_m^{i})^2+K\eta^3\frac{\sigma^2}{n}(n+1)\frac{\tau(\tau-1)}{2}L^2 \\
&+K\eta^2\frac{L}{2}\sigma^2\tau[\frac{1}{n}+\frac{1}{r(n-1)}(1-\frac{r}{n})4(1+q_s)n], \label{summed}
\end{split}
\end{equation}
or
\begin{align}
\frac{1}{K\tau}&\sum_{m=0}^{K-1}\sum_{t=0}^{\tau -1}\mathbb{E}\Vert \nabla f(\bar{X}_{m,t})\Vert^2 \leq \frac{2(f(X_0)-f^*)}{K\eta\tau} \nonumber \\
&+\frac{L}{Kn^2\eta\tau}\sum_{m=0}^{K-1}\sum_{i\in [n]}q_s( range_m^{i})^2+\eta^2\frac{\sigma^2}{n}(n+1)(\tau-1)L^2 \nonumber\\
&+\eta L\sigma^2[\frac{1}{n}+\frac{1}{r(n-1)}(1-\frac{r}{n})4(1+q_s)n].
\label{summed2}
\end{align}

In this paper and \cite{b12}, $r=n$, (\ref{summed2}) can be simplified into
\begin{align}
&\frac{1}{K\tau}\sum_{m=0}^{K-1}\sum_{t=0}^{\tau -1}\mathbb{E}\Vert \nabla f(\bar{X}_{m,t})\Vert^2 \leq \frac{Ld}{n^2\eta K\tau}\sum_{m=0}^{K-1}\sum_{i\in [n]}(\frac{range_m^{i}}{s_m^{i}})^2 \nonumber \\
&+\frac{2(f(X_0)-f^*)}{\eta K\tau}+\frac{\eta^2\sigma^2(n+1)(\tau-1)L^2}{n} +\frac{\eta \sigma^2L}{n}, 
\label{theorem}
\end{align}
which completes the proof of Theorem 1.

\bibliographystyle{IEEEbib}
\bibliography{strings,refs}

\begin{thebibliography}{10}

\bibitem{b1}
B.~McMahan, E.~Moore, D.~Ramage, S.~Hampson, and B.~A. y~Arcas,
\newblock ``Communication-efficient learning of deep networks from
  decentralized data,''
\newblock in {\em Artificial Intelligence and Statistics}. PMLR, 2017, pp.
  1273--1282.

\bibitem{b2}
P.~Kairouz, H.~B. McMahan, B.~Avent, A.~Bellet, M.~Bennis, A.~N. Bhagoji,
  K.~Bonawitz, Z.~Charles, G.~Cormode, R.~Cummings, et~al.,
\newblock ``Advances and open problems in federated learning,''
\newblock {\em arXiv preprint arXiv:1912.04977}, 2019.

\bibitem{b3}
N.~Guha, A.~Talwalkar, and V.~Smith,
\newblock ``One-shot federated learning,''
\newblock {\em arXiv preprint arXiv:1902.11175}, 2019.

\bibitem{b4}
J.~Mills, J.~Hu, and G.~Min,
\newblock ``{Communication-efficient federated learning for wireless edge
  intelligence in IoT},''
\newblock {\em IEEE Internet of Things Journal}, vol. 7, no. 7, pp. 5986--5994,
  2019.

\bibitem{b5}
A.~F. Aji and K.~Heafield,
\newblock ``Sparse communication for distributed gradient descent,''
\newblock {\em arXiv preprint arXiv:1704.05021}, 2017.

\bibitem{b6}
D.~Alistarh, D.~Grubic, J.~Li, R.~Tomioka, and M.~Vojnovic,
\newblock ``{QSGD: Communication-efficient SGD via gradient quantization and
  encoding},''
\newblock {\em Advances in Neural Information Processing Systems}, vol. 30, pp.
  1709--1720, 2017.

\bibitem{b7}
J.~Xu, W.~Du, Y.~Jin, W.~He, and R.~Cheng,
\newblock ``Ternary compression for communication-efficient federated
  learning,''
\newblock {\em IEEE Transactions on Neural Networks and Learning Systems},
  2020.

\bibitem{b8}
A.~{\O}land and B.~Raj,
\newblock ``Reducing communication overhead in distributed learning by an order
  of magnitude (almost),''
\newblock in {\em 2015 IEEE International Conference on Acoustics, Speech and
  Signal Processing (ICASSP)}. IEEE, 2015, pp. 2219--2223.

\bibitem{b9}
G.~Cui, J.~Xu, W.~Zeng, Y.~Lan, J.~Guo, and X.~Cheng,
\newblock ``Mqgrad: Reinforcement learning of gradient quantization in
  parameter server,''
\newblock in {\em Proceedings of the 2018 ACM SIGIR International Conference on
  Theory of Information Retrieval}, 2018, pp. 83--90.

\bibitem{b10}
J.~Bernstein, Y.-X. Wang, K.~Azizzadenesheli, and A.~Anandkumar,
\newblock ``{SignSGD: Compressed optimisation for non-convex problems},''
\newblock in {\em International Conference on Machine Learning}. PMLR, 2018,
  pp. 560--569.

\bibitem{b11}
J.~Guo, W.~Liu, W.~Wang, J.~Han, R.~Li, Y.~Lu, and S.~Hu,
\newblock ``Accelerating distributed deep learning by adaptive gradient
  quantization,''
\newblock in {\em ICASSP 2020-2020 IEEE International Conference on Acoustics,
  Speech and Signal Processing (ICASSP)}. IEEE, 2020, pp. 1603--1607.

\bibitem{b12}
D.~Jhunjhunwala, A.~Gadhikar, G.~Joshi, and Y.~C. Eldar,
\newblock ``Adaptive quantization of model updates for communication-efficient
  federated learning,''
\newblock in {\em ICASSP 2021-2021 IEEE International Conference on Acoustics,
  Speech and Signal Processing (ICASSP)}. IEEE, 2021, pp. 3110--3114.

\bibitem{b13}
A.~Reisizadeh, A.~Mokhtari, H.~Hassani, A.~Jadbabaie, and R.~Pedarsani,
\newblock ``{FedPAQ: A communication-efficient federated learning method with
  periodic averaging and quantization},''
\newblock in {\em International Conference on Artificial Intelligence and
  Statistics}. PMLR, 2020, pp. 2021--2031.

\bibitem{b14}
A.~Theertha~Suresh, F.~X. Yu, S.~Kumar, and H.~Brendan~McMahan,
\newblock ``Distributed mean estimation with limited communication,''
\newblock {\em arXiv e-prints}, pp. arXiv--1611, 2016.

\bibitem{b15}
H.~Xiao, K.~Rasul, and R.~Vollgraf,
\newblock ``{Fashion-MNIST: A novel image dataset for benchmarking machine
  learning algorithms},''
\newblock {\em arXiv preprint arXiv:1708.07747}, 2017.

\bibitem{b16}
A.~Krizhevsky, V.~Nair, and G.~Hinton,
\newblock ``{Cifar-10 (Canadian institute for advanced research)},''
\newblock {\em URL http://www. cs. toronto. edu/kriz/cifar. html}, vol. 5, pp.
  4, 2010.

\bibitem{b17}
K.~He, X.~Zhang, S.~Ren, and J.~Sun,
\newblock ``Deep residual learning for image recognition,''
\newblock in {\em Proceedings of the IEEE Conference on Computer Vision and
  Pattern Recognition}, 2016, pp. 770--778.

\bibitem{b18}
F.~Haddadpour, M.~M. Kamani, A.~Mokhtari, and M.~Mahdavi,
\newblock ``Federated learning with compression: Unified analysis and sharp
  guarantees,''
\newblock in {\em International Conference on Artificial Intelligence and
  Statistics}. PMLR, 2021, pp. 2350--2358.

\end{thebibliography}

\end{document}